\documentclass{article}

\usepackage{microtype}
\usepackage{graphicx}
\usepackage{subfigure}
\usepackage{booktabs} %

\usepackage{hyperref}

\usepackage[accepted]{icml2024}

\usepackage{amsmath}
\usepackage{amssymb}
\usepackage{mathtools}
\usepackage{amsthm}

\usepackage[capitalize,noabbrev]{cleveref}
\usepackage{makecell}
\usepackage[inline]{enumitem}
\usepackage{xspace}
\usepackage{tabularx}
\newcolumntype{C}{>{\centering\arraybackslash}X}

\makeatletter
\DeclareRobustCommand\onedot{\futurelet\@let@token\@onedot}
\def\@onedot{\ifx\@let@token.\else.\null\fi\xspace}
\makeatother

\newcommand{\eg}{\textit{e.g}\onedot}
\newcommand{\ie}{\textit{i.e}\onedot}

\newcommand{\vs}{\textit{vs.}\onedot}

\newcommand{\revise}[1]{{\color{black}#1}}
\theoremstyle{plain}

\theoremstyle{definition}

\theoremstyle{remark}

\makeatletter
\renewcommand{\paragraph}{%
  \@startsection{paragraph}{4}%
  {\z@}{0em}{-0.5em}%
  {\normalfont\normalsize\bfseries}%
}
\makeatother

\usepackage[textsize=tiny]{todonotes}

\icmltitlerunning{What Matters for Calibrating Vision--Language Models}

\begin{document}

\twocolumn[
\icmltitle{An Empirical Study Into What Matters for Calibrating Vision--Language Models}

\icmlsetsymbol{equal}{*}

\begin{icmlauthorlist}
\icmlauthor{Weijie Tu}{anu}
\icmlauthor{Weijian Deng}{anu}
\icmlauthor{Dylan Campbell}{anu}
\icmlauthor{Stephen Gould}{anu}
\icmlauthor{Tom Gedeon}{anu,curtin,obuda}
\end{icmlauthorlist}

\icmlaffiliation{anu}{The Australian National University}
\icmlaffiliation{curtin}{Curtin University}
\icmlaffiliation{obuda}{University of \'Obuda}

\icmlcorrespondingauthor{Weijie Tu}{weijie.tu@anu.edu.au}

\icmlkeywords{Machine Learning, ICML}

\vskip 0.3in
]

\printAffiliationsAndNotice{}  %

\begin{abstract}
Vision--Language Models (VLMs) have emerged as the dominant approach for zero-shot recognition, adept at handling diverse scenarios and significant distribution changes. However, their deployment in risk-sensitive areas requires a deep understanding of their uncertainty estimation capabilities, a relatively uncharted area. In this study, we explore the calibration properties of VLMs across different architectures, datasets, and training strategies. In particular, we analyze the uncertainty estimation performance of VLMs when calibrated in one domain, label set or hierarchy level, and tested in a different one. Our findings reveal that while VLMs are not inherently calibrated for uncertainty, temperature scaling significantly and consistently improves calibration, even across shifts in distribution and changes in label set. Moreover, VLMs can be calibrated with a very small set of examples. Through detailed experimentation, we highlight the potential applications and importance of our insights, aiming for more reliable and effective use of VLMs in critical, real-world scenarios.
\end{abstract}

\section{Introduction}
\label{sec:intro}
Vision--language models (VLMs), such as CLIP~\cite{radford2021learning} and ALIGN~\cite{jia2021scaling}, have achieved remarkable results for a wide range of tasks, such as zero-shot image recognition~\cite{wortsman2022robust}, open-vocabulary object detection~\cite{zhou2022detecting, gu2021open}, image captioning~\cite{yu2022coca, mokady2021clipcap} and egocentric perception~\cite{zeng2022socratic}. 
The burgeoning field of VLMs has been characterized by rapid exploration along various dimensions~\cite{nguyen2022quality, fang2022data, wortsman2022robust,cherti2022reproducible,tu2023closer}, such as dataset creation~\cite{nguyen2022quality}, reproducible scaling laws~\cite{cherti2022reproducible}, compositional relationships between objects and attributes~\cite{yuksekgonul2022and}, robust fine-tuning approaches~\cite{goyal2023finetune}, and visual factor-level robustness~\cite{tu2023closer}.

However, their application in risk-sensitive domains necessitates a more rigorous understanding of their uncertainty estimation capabilities, an area that remains largely under-explored.
Model \textit{calibration} is concerned with ensuring that the model's predicted output probabilities correspond to its empirical frequency of correctness (\ie, accuracy).
For example, a calibrated model that classifies some images as ``cow'' with a 50\% probability will have misclassified roughly half. 
\citet{galil2023can} and \citet{minderer2021revisiting} report that CLIP models are better calibrated than other models trained on ImageNet. 
Notwithstanding this observation, \citet{tu2023closer} point out that they are not always well-calibrated and attribute this to the impact of the training data distribution and quantity.
Building on this line of research, we study the calibration properties of various VLMs, each characterized by different architectures, datasets, and training strategies.

We investigate which factors affect the calibration of VLMs.
Starting from prior research that demonstrates zero-shot CLIP can be well-calibrated with simple temperature scaling under distribution shifts~\cite{tu2023closer}, we extend this analysis to other CLIP variants and exemplar vision--language models. 
We then examine whether such a property persists when the calibration dataset varies in (1) distribution, (2) label set (\eg, CIFAR-10 \vs ImageNet), (3) hierarchy level (\eg, ``\textit{Spider}'' \vs ``\textit{Black widow}''), (4) the number of images, and (5) feature-space distance of calibration set with respect to the target test set.

To this end, we evaluate $35$ vision--language models. They have various image--text pre-training frameworks, such as CLIP \cite{radford2021learning} and BLIP \cite{li2022blip}. They also have different visual encoder architectures (\eg, ViT \cite{shankar2021image} and ConvNeXt \cite{liu2022convnet}) and training dataset distributions and quantities. We assay the uncertainty estimation of VLMs on three standard image classification benchmarks: ImageNet \cite{imagenet_cvpr09}, CIFAR-10 \cite{krizhevsky2009learning} and DomainNet \cite{peng2019moment} and $5$ types of distribution shift, including reproduction shift \cite{recht2019imagenet} and sketch shift \cite{wang2019learning}. Moreover, to study the sensitivity of our findings, we analyze the uncertainty estimation performance of VLMs when the quantity and quality of the calibration set are varied and the text prompts are hand-crafted or machine-generated. 
Our key observations are:
\begin{itemize}[topsep=-2pt,itemsep=-0.1ex]
  \item 
  {After calibrating all models with temperature scaling~\cite{platt1999probabilistic}, vision--language models are better calibrated than the other models in this study, which is not necessarily the case prior to calibration;}
  \item VLMs can be calibrated on datasets with different label sets than the target set and can be calibrated at a higher or lower level of the label hierarchy than the level of the target labels;
  \item VLMs require a few samples for calibration. \revise{For example, VLMs can be calibrated using temperature scaling, spline fitting \cite{gupta2021calibration} or histogram binning \cite{zadrozny2001obtaining} with less than $100$ samples};
  \item VLMs do not require sophisticated prompting strategies for calibration, with ``\textit{a photo of a $<$class$>$}'' being sufficient to achieve good uncertainty estimation;
  \item Our findings motivate the use of a synthetic calibration set for VLMs in practical settings where labeled calibration data is lacking.
\end{itemize}

\section{Related Work}
\label{sec:related}

\paragraph{Vision--language models} have demonstrated strong capabilities by leveraging web-scale datasets and language supervision to learn joint image--text representations \cite{bommasani2021opportunities, radford2021learning, jia2021scaling}. A seminal work by \citet{radford2021learning} introduced CLIP, a large VLM trained on $400$ million filtered web-crawled image--text pairs, which exhibits unprecedented zero-shot ability on numerous downstream visual tasks. Inspired by CLIP, various algorithms have been created to enhance the performance of the model \cite{singh2022flava, li2022blip, zhai2023sigmoid, li2023blip}. For example, \citet{li2022blip} propose BLIP, a new pre-training framework, that bootstraps captions to effectively leverage noisy web data. \citet{zhai2023sigmoid} designs a simple pairwise sigmoid loss which solely operates on the image--text pairs without requiring the global view of pairwise similarities for normalization. 

Encouraged by the strong generalizability of VLMs, researchers have explored their properties from diverse perspectives, such as robustness and bias. For instance, \citet{schiappa2022multi} and \citet{qiu2022multimodal} investigate their robustness through perturbations, while \citet{fang2022data} attribute the remarkable robustness of CLIP to its diverse training distribution. Additionally, \citet{yuksekgonul2022and} and \citet{thrush2022winoground} assess the capability of VLMs to encode compositional information. \citet{liang2022mind} study the modality gap from the perspectives of model initialization and contrastive learning optimization.
While~\citet{tu2023closer} focus on the calibration of CLIP models using temperature scaling on ImageNet, our study extends this to a diverse array of VLMs and various calibration methods (\textit{e.g.}, Spline, histogram binning, and vector scaling). We provide an in-depth analysis of calibration factors, examine the uncertainty estimates of VLMs, and consider different aspects that may influence calibration performance. Furthermore, we demonstrate the practical utility of our findings in a real-world problem setup.

\paragraph{{Confidence calibration}} aims to calibrate models so that their prediction probabilities align with the empirical frequency of correctness~\cite{nguyen2015posterior, guo2017calibration}. Much research effort has been made in proposing algorithms to improve model calibration performance, such as post-hoc rescaling the prediction probabilities \cite{guo2017calibration}, ensembling \cite{lakshminarayanan2017simple} and pre-training \cite{hendrycks2019using}. Another line of research focuses on analyzing calibration of modern neural networks \cite{guo2017calibration, ovadia2019can, minderer2021revisiting, tu2023closer}. \citet{guo2017calibration} point out that modern neural networks are poorly calibrated. \citet{ovadia2019can} observe that distribution shifts degrade the performance of calibration methods. 
\citet{minderer2021revisiting} show that zero-shot CLIP models are well-calibrated given their performance. 
\citet{tu2023closer} show that zero-shot CLIP models are well-calibrated with temperature scaling.
This paper builds on this prior research by studying a more comprehensive suite of factors that influence the uncertainty estimation performance of VLMs.

\begin{figure*}[t]
  \begin{center}
    \includegraphics[width=\linewidth]{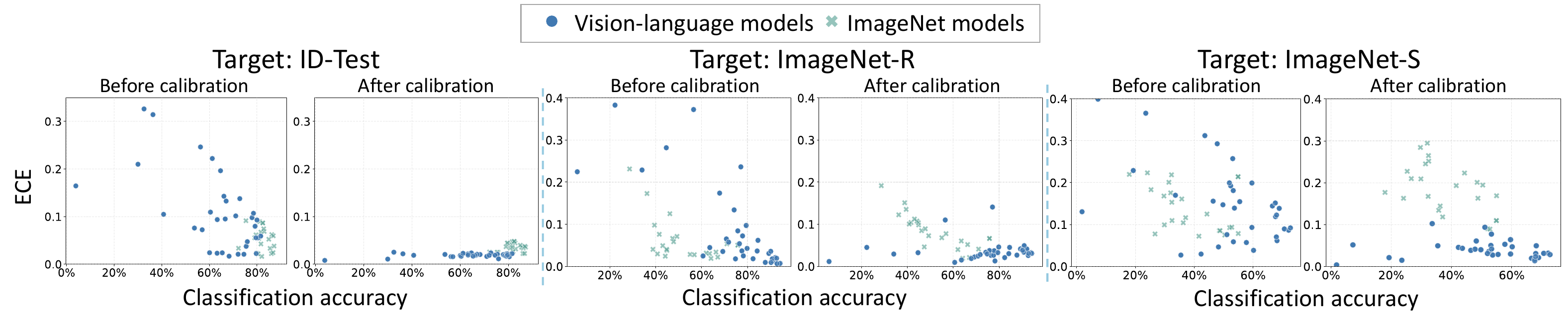}
      \vspace{-20pt}
    \caption{\textbf{Comparing the calibration performance of ImageNet-trained models and VLMs}. We report the results on the in-distribution test set (ID-Test) and two out-of-distribution (OOD) test sets: ImageNet-R and ImageNet-S. We plot the expected calibration error (ECE) before and after temperature scaling for each model. The blue dots represent VLMs and the green crosses denote ImageNet-trained models. We observe that VLMs are well-calibrated by temperature scaling on both ID and OOD test sets.
    }\label{fig:comparison}
  \end{center}
  \vspace{-10pt}
\end{figure*}

\section{Definition and Notation} 
Let $\mathcal{Y} = {1, \ldots, K}$ and $\mathcal{X} = \mathbb{R}^d$ denote label and input spaces, respectively. A sample $(\mathbf{x}, y)$ from an unknown distribution in $\mathcal{X} \times \mathcal{Y}$ is input to a neural network classifier $\mathbf{f}: \mathcal{X} \rightarrow \Delta^k$. This classifier outputs a probability distribution over $k$ classes for $\mathbf{x}$, where $\Delta^k$ is the $k - 1$ dimensional simplex.
We assume $\mathbf{f}$ combines two functions: $\mathbf{f} \eqqcolon \mathbf{\sigma} \circ \mathbf{g}$, where $\mathbf{g}: \mathbb{R}^d \to \mathbb{R}^k$ is a non-probabilistic $k$-way classifier, and $\mathbf{\sigma}: \mathbb{R}^k \to \Delta^k$ is the softmax operator $\mathbf{\sigma}_i(\mathbf{z}) = \frac{\exp(\mathbf{z}_i)}{\sum_{j=1}^k \exp(\mathbf{z}_j)}$ for $i \in \mathcal{Y}$.
The output $\mathbf{g}(\mathbf{x})$ is referred to as the logits of $\mathbf{x}$ relative to $\mathbf{f}$.
For any input instance $\mathbf{x}$, $\mathbf{f}$ assigns the predicted label $\hat{y} \eqqcolon \arg\max_{i} \mathbf{f}_{i}(\mathbf{x})$ and the corresponding confidence score $\hat{p} \eqqcolon \max_i \mathbf{f}{i} (\mathbf{x})$.

\paragraph{Expected Calibration Error (ECE).} 
A model is perfectly calibrated if $\mathbb{P}(\hat{y}=y\mid\hat{p}=p)=p$ for all $p$ in [0,1], where~$y$ is the actual label, $\hat{y}$ the prediction, and $\hat{p}$ the confidence score. To assess model calibration, we typically use the Expected Calibration Error (ECE) \cite{guo2017calibration}, lower values indicating better calibration. ECE involves dividing samples into $M$ equal bins by confidence scores, then computing the mean absolute difference between each bin's accuracy and average confidence: $\mathrm{ECE} =\sum_{m=1}^M\frac{|B_m|}{n}|\text{acc}(B_m)-\text{avgConf}(B_m)|$, with $n$ as the total number of samples

\paragraph{Temperature Scaling.} 
Scaling logits from $\mathbf{g}$ with temperature $T$ modifies output probability sharpness. The new prediction confidence is $\hat{p} =\max_i \frac{\exp(\mathbf{g}_i(\mathbf{x})/T)}{\sum_{j=1}^n \exp(\mathbf{g}_j(\mathbf{x})/T)}$. Higher $T$ softens, and lower $T$ sharpens probabilities. As $T$ approaches 0 or infinity, probabilities trend towards a one-hot vector or uniform distribution, respectively. For a trained classifier $\mathbf{f}$, $T$ is optimized using negative log-likelihood~(NLL) on a calibration set. Since $T$ does not impact the softmax maximum, $\hat{y}$, the predicted class, remains the same, preserving classification accuracy.

\section{Experimental Setup}
\label{sec:setup}

\paragraph{Compared models: VLMs.}
We consider $35$ vision--language models. These models consist of exemplar VLMs, such as CLIP~\cite{radford2021learning}, Flava \cite{singh2022flava} and BLIP \cite{li2022blip}. In particular, we evaluate zero-shot CLIP models that are trained on different training distributions, such as the WIT \cite{radford2021learning} and LAION \cite{gadre2023datacomp} datasets, diverse dataset quantities from $3$ million to $2$ billion, and curated training datasets \cite{xu2023demystifying}. CLIP variants with different image encoders are also assessed, including ViT \cite{dosovitskiy2020image} and ConvNeXt \cite{liu2022convnet} encoders, as well as CLIP variants that modify the training objective, such as SigLIP \cite{zhai2023sigmoid}, or the training strategy \cite{li2023inverse}. Unless specified, for each model, we use their default prompt sets from \citet{radford2021learning}.

\paragraph{Compared models: non-VLMs.}
We compare VLM calibration performance with models trained on ImageNet to show that VLMs are well-calibrated despite the distribution shift after temperature scaling. We consider convolutional neural networks, such as ResNet \cite{he2016deep} and ConvNeXt \cite{liu2022convnet}, and vision transformers, exemplified by ViT \cite{dosovitskiy2020image} and Swin \cite{liu2021swin}. These models are trained solely on ImageNet \cite{imagenet_cvpr09} or pre-trained on a significantly larger dataset (\eg, ImageNet-21K \cite{ridnik2021imagenet21k}). All the modelsmentioned are publicly accessible through OpenCLIP \cite{ilharco_gabriel_2021_5143773} and TIMM \cite{rw2019timm}. 

\paragraph{Test sets.}
We evaluate the calibration of VLM on three standard image classification benchmarks: ImageNet \cite{imagenet_cvpr09}, CIFAR-10 \cite{krizhevsky2009learning} and DomainNet \cite{peng2019moment}. 
Following the protocol in \cite{gupta2021calibration}, we divide the validation set of ImageNet into two halves: one for the in-distribution (ID) test set, and the other for learning calibration methods.
OOD test sets are ImageNet-V2-A \cite{recht2019imagenet}, ImageNet-R(endition) \cite{hendrycks2021many}, ImageNet-S(ketch) \cite{wang2019learning}, and ObjectNet \cite{barbu2019objectnet}. For CIFAR-10, its validation set is used for model calibration, and CIFAR-10.1, CIFAR-10.2 \cite{recht2018cifar10.1} and CINIC \cite{darlow2018cinic} are used for evaluation. The DomainNet benchmark utilizes the `Real' domain for calibration and evaluates on `Painting' and `Sketch' domains. Note that ImageNet-R and ObjectNet use a reduced subset of classes; we follow the literature \cite{bello2021revisiting} to select subset of logits for these classes before evaluation.

\begin{figure*}[t]
    \centering
    \includegraphics[width=\linewidth]{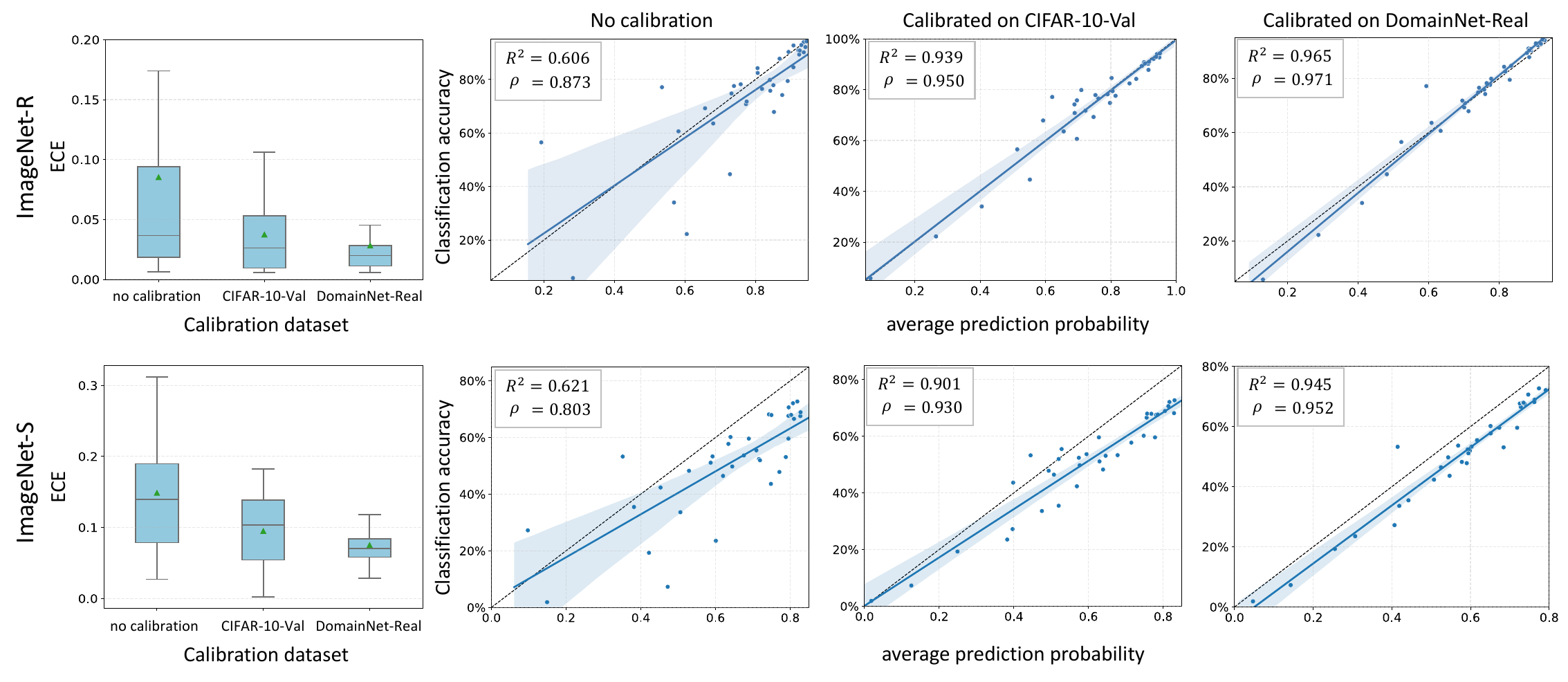}
    \vspace{-15pt}
    \caption{
    \textbf{Adaptability of VLMs to different calibration label sets.}
    \textbf{Left: Calibration error reduction.}
    Here, we observe a significant decrease in the expected calibration error for VLMs following cross-label-set calibration, as opposed to when no calibration is applied.
    \textbf{Right: Correlation between VLM prediction probability and classification accuracy.}
    This graph illustrates the classification accuracy of VLMs on ImageNet-R and ImageNet-S against their average prediction probability, before and after calibration with CIFAR-10-Val or DomainNet-Real. Each point represents a model, with the dashed black line indicating perfect calibration (y=x). The data showcases a strong linear and rank correlation, even when models are calibrated on label sets different from the target, proving the effectiveness of cross-label-set calibration for VLMs.
    }
    \label{fig:crosstask}
\end{figure*}

\paragraph{Calibration method.}
For model calibration, we by default use temperature scaling~\cite{guo2017calibration} on calibration sets. We also experiment with the spline post-hoc calibration method \cite{gupta2021calibration}.

\paragraph{Metrics.} 

(1)~\underline{Calibration metric:} we use ECE as the evaluation metric, where a lower score indicates better calibration performance. Throughout the experiments, we estimate ECE using equal-mass binning and $15$ bins.
(2)~\underline{Correlation metric:} to examine whether the calibrated prediction probabilities for all models exhibit a correlation with their classification accuracy, we use coefficients of determination $R^2$ \cite{nagelkerke1991note} to measure the linearity and utilize Spearman's rank coefficient $\rho$ \cite{kendall1948rank} to measure monotonicity. $R^2$ ranges from $0$ to $1$, where an $R^2$ of $1$ means that regression predictions perfectly correlate with model performance. The rank coefficient $\rho$ spans $[-1, 1]$, where a value closer to $1$ (or $-1$) indicates a better ranking index, while $0$ indicates no correlation.

\section{Factors Affecting Calibration}
\label{sec:calibration}

The cornerstone of safely deploying Vision--Language Models (VLMs) lies in verifying their decision reliability.
Specifically, the prediction probabilities provided by VLMs should accurately reflect their performance. With this objective, we have embarked on a comprehensive set of experiments. These are designed to scrutinize the uncertainty estimation of VLMs under various conditions, including changes in (1) distribution, (2) label sets (e.g., CIFAR-10 vs. ImageNet), (3) hierarchy levels (e.g., ``Spider'' vs. ``Black widow''), (4) the number of images in the dataset, and (5) the feature-space distance between the calibration and target test sets.

\paragraph{VLMs are well-calibrated after temperature scaling across various distributions.}
\cref{fig:comparison} compares the calibration performance of VLMs with models trained on ImageNet. On both ID and OOD test sets (ImageNet-S and ImageNet-R), we observe that before calibration with temperature scaling, VLMs do not necessarily have superior uncertainty estimation performance. For instance, certain ImageNet models demonstrate a lower Expected Calibration Error (ECE) before calibration. However, once temperature scaling is applied, the scenario changes dramatically. VLMs show a marked decrease in their average ECE, dropping to $0.05$, whereas ImageNet models see an increase in ECE, rising to $0.15$. This indicates that VLMs benefit more significantly from the calibration process than ImageNet models.

Furthermore, while existing research highlights the challenges in achieving stable calibration results under distribution shifts \cite{yu2022robust,Zou_2023_ICCV,tomani2023beyond,ovadia2019can}, VLMs manage to maintain consistent and reliable uncertainty estimation after temperature scaling. This is evident in their performance on OOD test sets (ImageNet-S and ImageNet-R), where they exhibit competent uncertainty estimation. This enhanced calibration capability of VLMs, especially after temperature scaling, underscores their potential for more accurate and dependable decision-making in diverse applications.

\subsection{Adaptability to Different Calibration Label Sets}
\label{sec:crosstask}
\paragraph{VLMs can be calibrated on a dataset with a different label set from the target dataset.}
The zero-shot capability of VLMs facilitates their direct application to a diverse array of downstream classification tasks without the need for explicit training or fine-tuning. The conventional boundary between ID and OOD classes becomes less clear-cut for VLMs, suggesting its potential for cross-label-set calibration. 
To evaluate the effect of calibration label sets, we conduct experiments calibrating VLMs on datasets with different label sets. The findings, depicted in \cref{fig:crosstask}, demonstrate an improvement in the uncertainty estimation of VLMs when calibrated with alternative label sets. For instance, calibrating VLMs on CIFAR-10-Val or DomainNet-Real significantly reduces the ECE on ImageNet-R compared to when no calibration is applied. This trend of reduced ECE is consistently observed on ImageNet-S as well, further validating the effectiveness of the cross-label-set calibration.
Furthermore, the calibrated prediction probability strongly correlates with model accuracy, with a linear and rank correlation over $0.90$, despite the presence of non-zero ECE.
This indicates that, even with label set differences, the calibrated prediction probability is predictive of the rankings of VLMs.

\begin{figure}\centering
\includegraphics[width=\linewidth]{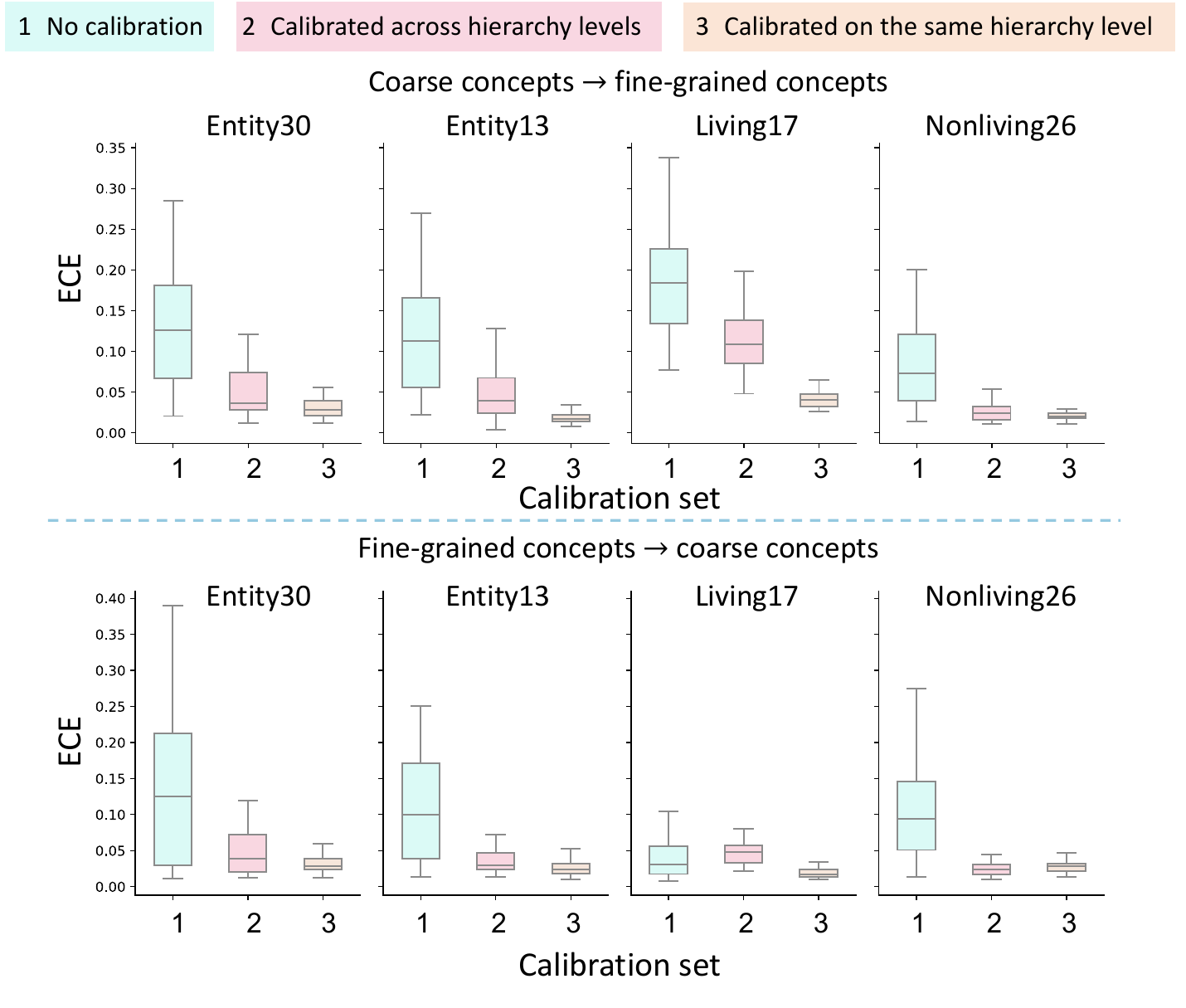}
\vspace{-20pt}
\caption{
\textbf{Robustness of VLM calibration to label hierarchy levels.}
This figure presents box plots summarizing the calibration errors (ECEs) of VLMs calibrated with label hierarchies differing in granularity from the target dataset (ImageNet-S). The top row shows calibration at a coarser level, and the bottom row at a finer level. Despite not matching the calibration precision of same-level calibration, the minimal differences indicate the robustness of VLM calibration to label granularity.
}
\label{fig:breeds}
\vspace{-10pt}
\end{figure}

\subsection{Calibration Across Semantic Hierarchy Levels}
\label{sec:hierarchy}

VLMs perform zero-shot classification by generating query embeddings for each novel class from their natural language names.
Label hierarchy sets have been shown effective in enhancing model accuracy~\cite{novack2023chils,ren2023chatgpt}. For example, mapping the predicted sub-class back to its parent to produce the final prediction~\cite{novack2023chils}.
However, there has been little attention on the influence of a dataset's label hierarchy for calibration.
This section aims to investigate whether VLMs can be effectively calibrated across different levels of a semantic hierarchy. Specifically, we examine whether VLMs can be calibrated using a dataset with coarsely-defined concepts (\eg, ``\textit{Bag}'') while the target dataset comprises fine-grained concepts (\eg, ``\textit{Backpack}''), and vice versa. 

To conduct our evaluation, we use four label sets from BREEDS \cite{santurkar2020breeds}---Entity13, Entity30, Living17, and Nonliving26---that define a hierarchical mapping between coarse and fine-grained classes. For each set, we adhere to the associated hierarchical mapping to selectively curate and relabel images sourced from the ImageNet validation set. This process yields a \textit{calibration} dataset featuring coarse concepts. Subsequently, we curate the corresponding fine-grained class subset from ImageNet-S, thereby establishing a distinct \textit{target} test dataset with a different distribution.
We then repeat this procedure in reverse to calibrate with fine-grained concepts and test on coarse concepts.

\paragraph{VLMs can be calibrated across label hierarchy levels.}
\cref{fig:breeds} plots the expected calibration errors of VLMs calibrated at a different label hierarchy level than the target dataset.
We see that for all four sets, the ECE decreases considerably when calibrated at the `wrong' level of the hierarchy.
While not as well-calibrated as the models where the calibration and target sets were at the same level of the hierarchy, the difference is not substantial, especially for `Nonliving26'.
This suggests that calibration is relatively robust to the granularity of the labels in the calibration set.

\begin{figure}\centering
    \includegraphics[width=\linewidth]{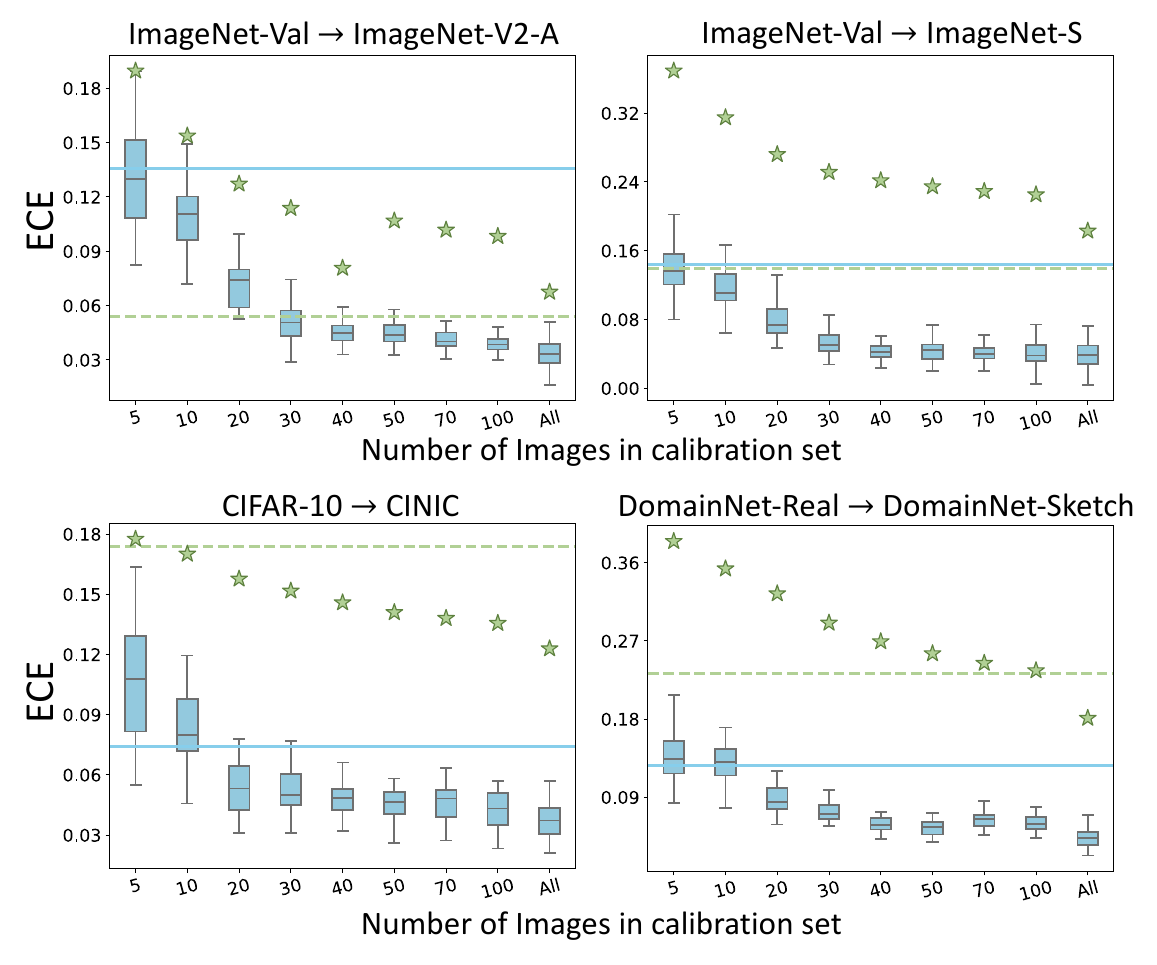}
      \vspace{-25pt}
    \caption{
    \textbf{Data-efficiency of VLM calibration across diverse datasets.}
        This figure displays the ECE of VLMs as a function of the calibration set size across four datasets: ImageNet-V2-A, ImageNet-S, CINIC, and DomainNet. {The green stars are the average ECE of calibrated models trained on the dataset. The blue \revise{solid} and green \revise{dashed} horizontal lines represent the average ECE before calibration of VLMs and non-VLMs, respectively.}
        The ECE values, averaged over ten random seeds, plateau after including merely $40$–$50$ images in the calibration set, chosen at random, 10 times. The results closely approximate the error obtained using the full set. This trend is observed despite the high number of classes in DomainNet and ImageNet, where many classes may not be represented even in the calibration set. These results highlight the data-efficiency of VLM calibration.
    }\label{fig:size}
    \vspace{-18pt}
\end{figure}

\subsection{Effect of the Calibration Set Size}
\label{sec:size}
In this section, we investigate whether VLMs demand a large number of images for calibration.
We evaluate VLMs on ImageNet-V2-A, ImageNet-S, CINIC and DomainNet. For each target test set, we randomly sample a certain number of images from the corresponding calibration set as described in \cref{sec:setup}. The performance for each model is measured by the averaged ECE over ten random seeds.

\paragraph{VLMs can be calibrated with a very small number of images.}
\cref{fig:size} plots the calibration error of VLMs with respect to the size of the calibration set for four different target datasets. 
The calibration error plateaus after $40$-$50$ images and is already close to the error after calibration with the entire set.
Notably, DomainNet and ImageNet are $345$-way and $1000$-way classification tasks, and so many classes do not have a single image present in the calibration set.
This suggests that calibrating VLMs is very data-efficient, and that it is a low-dimensional problem.

\begin{figure}
  \begin{center}
    \includegraphics[width=\linewidth]{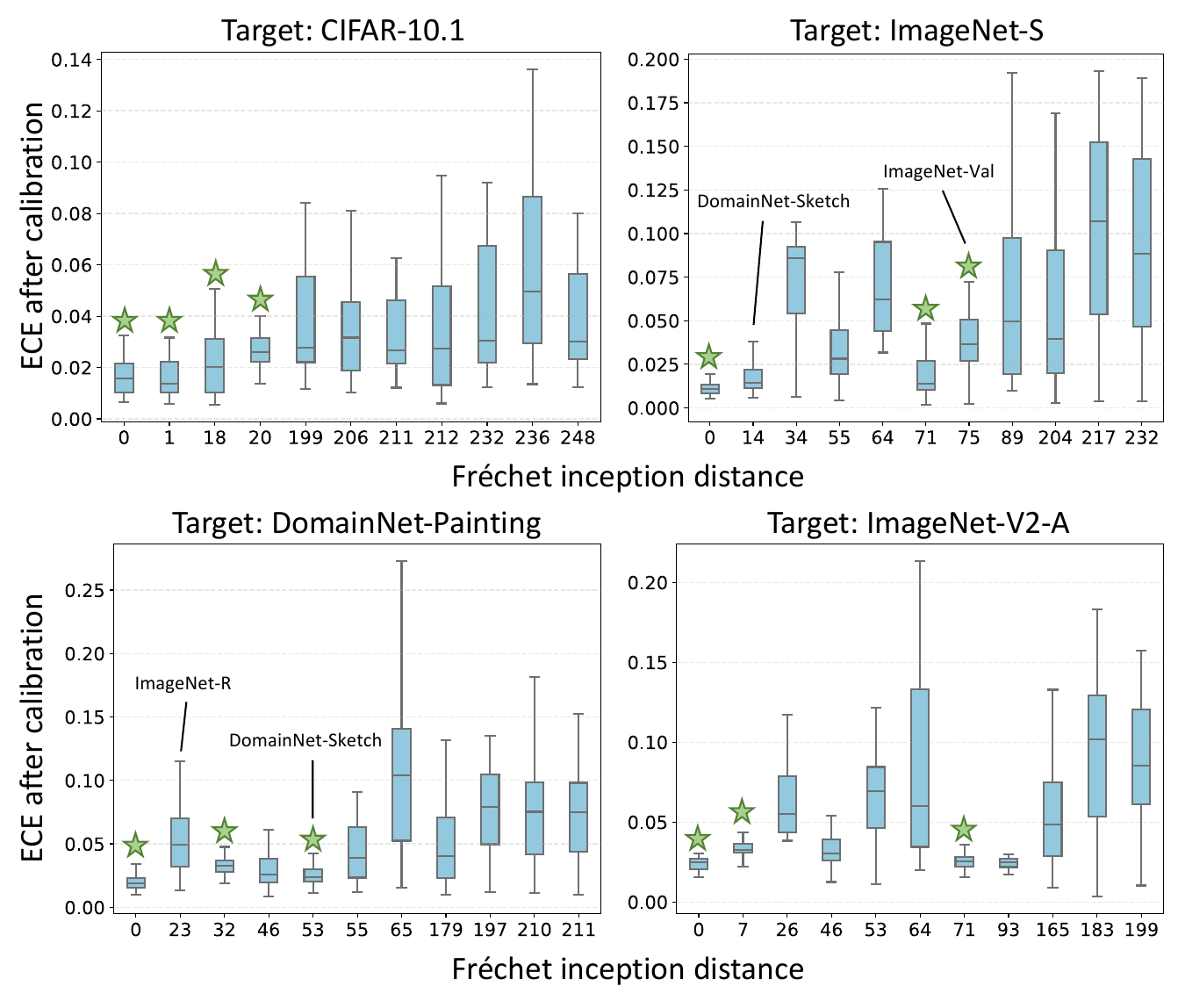}
      \vspace{-20pt}
    \caption{
    \textbf{Impact of the distance between calibration and target set distributions on VLM uncertainty estimates.}
    The distance between the calibration dataset and target dataset is computed by Fr\'echet inception distance \cite{heusel2017gans}.
    A green star indicates that the dataset for this column has the same label set as the target dataset.
    We find that the calibration error has a weak correlation with the FID between the calibration and target datasets, however the label set compatibility also plays a significant role.
    }\label{fig:distance}
  \end{center}
  \vspace{-15pt}
\end{figure}

\subsection{Effect of Calibration-Target Set Distance}
\label{sec:distance}
\citet{ovadia2019can} report that uncertainty estimation performance consistently degrades when calibration and test sets have distribution shifts.
In this section, we explore whether the same applies to VLMs.
Note that, since we have demonstrated that VLMs can be calibrated on a different label set, this experiment compares the calibration performance of VLMs calibrated on a dataset with the same or different label sets. The distribution discrepancy of datasets is computed by Fr\'echet inception distance (FID) \cite{heusel2017gans}. A zero FID means that the calibration set and target dataset follow a similar distribution.

\cref{fig:distance} plots the calibration error of VLMs with respect to the FID between the calibration and target datasets for four target datasets: CIFAR-10.1, ImageNet-V2-A, ImageNet-S, and DomainNet-Painting. We find that FID alone does not entirely explain the calibration error. Instead, it is a combination of the FID and the label set similarity that is more predictive of calibration performance.
This observation suggests that we should consider both the distribution shift and label set differences when selecting a calibration set.

\begin{figure}\centering
    \includegraphics[width=\linewidth]{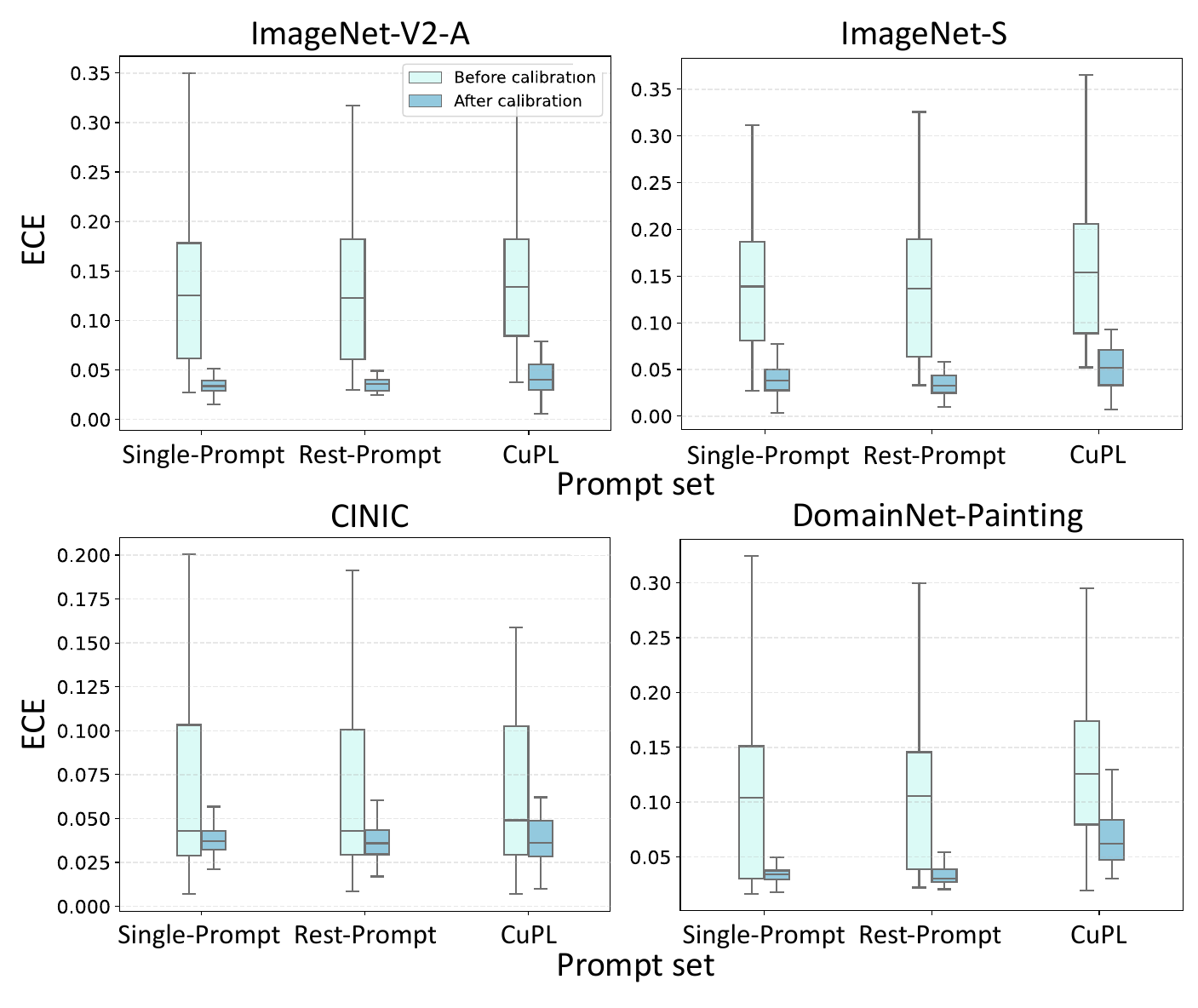}
      \vspace{-20pt}
    \caption{
    \textbf{Efficient calibration with single prompts for VLMs.}
    The figure illustrates the transferability of calibration effectiveness when using a single prompt for VLMs. The temperature scalar estimated using one random prompt, denoted as ``Single-Prompt" remains highly effective when applied to other human-designed prompts (``Rest-Prompt'') and machine-generated prompts (``CuPL''). This finding highlights the efficiency of single prompts in achieving robust calibration for VLMs.
    }\label{fig:prompt}
  \vspace{-5pt}
\end{figure}
\begin{figure*}[t]
    \centering
    \includegraphics[width=\linewidth]{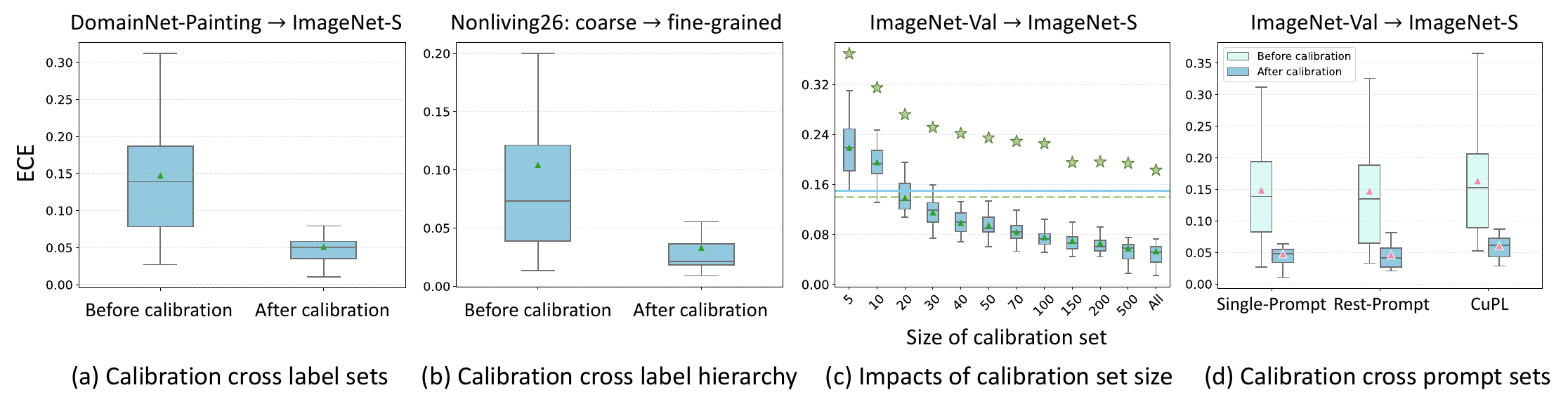}
    \vspace{-20pt}
    \caption{
   \textbf{Uncertainty estimation performance of VLMs with spline calibration.}
    Performance evaluation of VLMs using the spline method~\cite{gupta2021calibration} across four factors: (a) label sets, (b) label hierarchy, (c) calibration set size, and (d) prompt sets. Our results demonstrate that the phenomena observed with temperature scaling persist when employing spline calibration. For example, VLMs can be effectively calibrated on datasets with differing label sets or label hierarchies compared to the target test set.
    }
    \label{fig:spline}
    \vspace{-5pt}
\end{figure*}

\subsection{Transferability of Calibration Across Prompts}
\label{sec:prompt}
VLMs classify images by comparing image features with class weights computed by the text encoder, which takes as input the textual prompts describing each class of interest. This suggests that given the same textual class names, a better prompt set may improve a VLM's discriminative ability, and using specific prompt contexts for the style of images may also enhance model performance \cite{pratt2023does, zhou2022learning}. In the previous experiment, we use the same set of prompts for classification and calibration.
Here, we question whether a temperature scalar determined by one set of prompts can effectively calibrate a different set of prompts used for classification.

To answer this question, we conducted experiments where we randomly selected a single prompt from a larger set for calibration. Surprisingly, we found that calibrating with just a single prompt yielded effective results. The temperature scalar discovered through this single prompt calibration extended its effectiveness to both the rest of the human-designed prompts (Rest-Prompt) and machine-generated prompts (CuPL), as depicted in \cref{fig:prompt}.
This discovery highlights a key insight: VLMs do not necessarily require complex and tailored prompting strategies for calibration. Instead, a straightforward prompt such as ``\textit{a photo of a $<$class$>$}'' suffices for effective calibration.

\begin{figure}\centering
    \includegraphics[width=\linewidth]{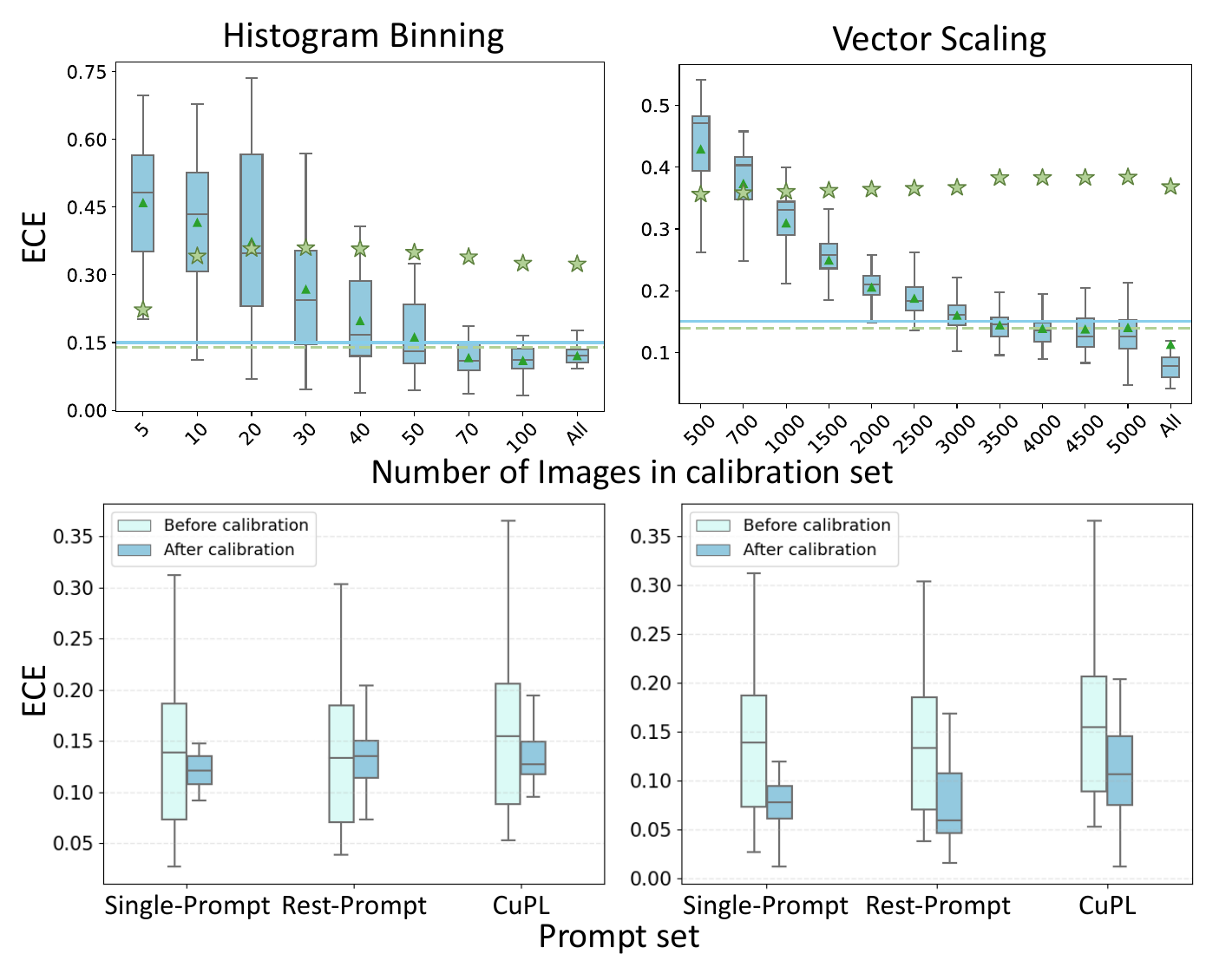}
      \vspace{-25pt}
    \caption{ \revise{
    \textbf{Uncertainty estimation performance of VLMs with histogram binning and vector scaling} across two factors: calibration set size (up) and prompt sets (bottom). The target domain is ImageNet-S and the calibration set is ImageNet-Validation.}
    }\label{fig:others}
  \vspace{-15pt}
\end{figure}

\subsection{Impact of the Calibration Method}
\label{sec:methods}

In addition to temperature scaling, we undertake an evaluation of three post-hoc calibration methods: spline fitting~\cite{gupta2021calibration}, histogram binning \cite{zadrozny2001obtaining} and vector scaling \cite{guo2017calibration}. 
Spline calibration involves deriving a function through spline fitting (SF), directly aligning classifier outputs with calibrated probabilities. Histogram binning (HB) and vector scaling (VS) both use class-specific parameters and assume an identical label set between calibration set and target set. Hence, HB and VS do not support calibration across label sets and label hierarchy levels.
Additionally, as suggested by \citep{guo2017calibration}, VS requires the presence of samples for \textit{each} class, meaning that VS demands more images for calibration.
This section investigates whether the observed calibration properties of VLMs hold using SF, HB and VS. 

In \cref{fig:spline} and \cref{fig:others}, we present our findings on SF, HB, and VS calibration methods. First, we observe that VLMs achieve lower ECE than calibrated non-VLMs using the three calibration methods under a distribution shift. Second, both SF and HB effectively calibrate VLMs using a small number of samples (\textit{\ie} $100$). In contrast, VS requires a larger number of images ($10\%$ of $50,000$ samples) for effective calibration, consistent with the findings in \citet{guo2017calibration}. Third, similar to temperature scaling, both SF and VS can calibrate VLMs using a single prompt, which can differ from the test prompt set. While HB effectively reduces the overall average ECE, it may not benefit all outliers. Lastly, when calibrating VLMs using datasets with different label sets or hierarchy levels compared to the target test set, spline calibration exhibits trends and outcomes similar to those of temperature scaling.

\begin{figure*}[t]
    \centering
    \includegraphics[width=\linewidth]{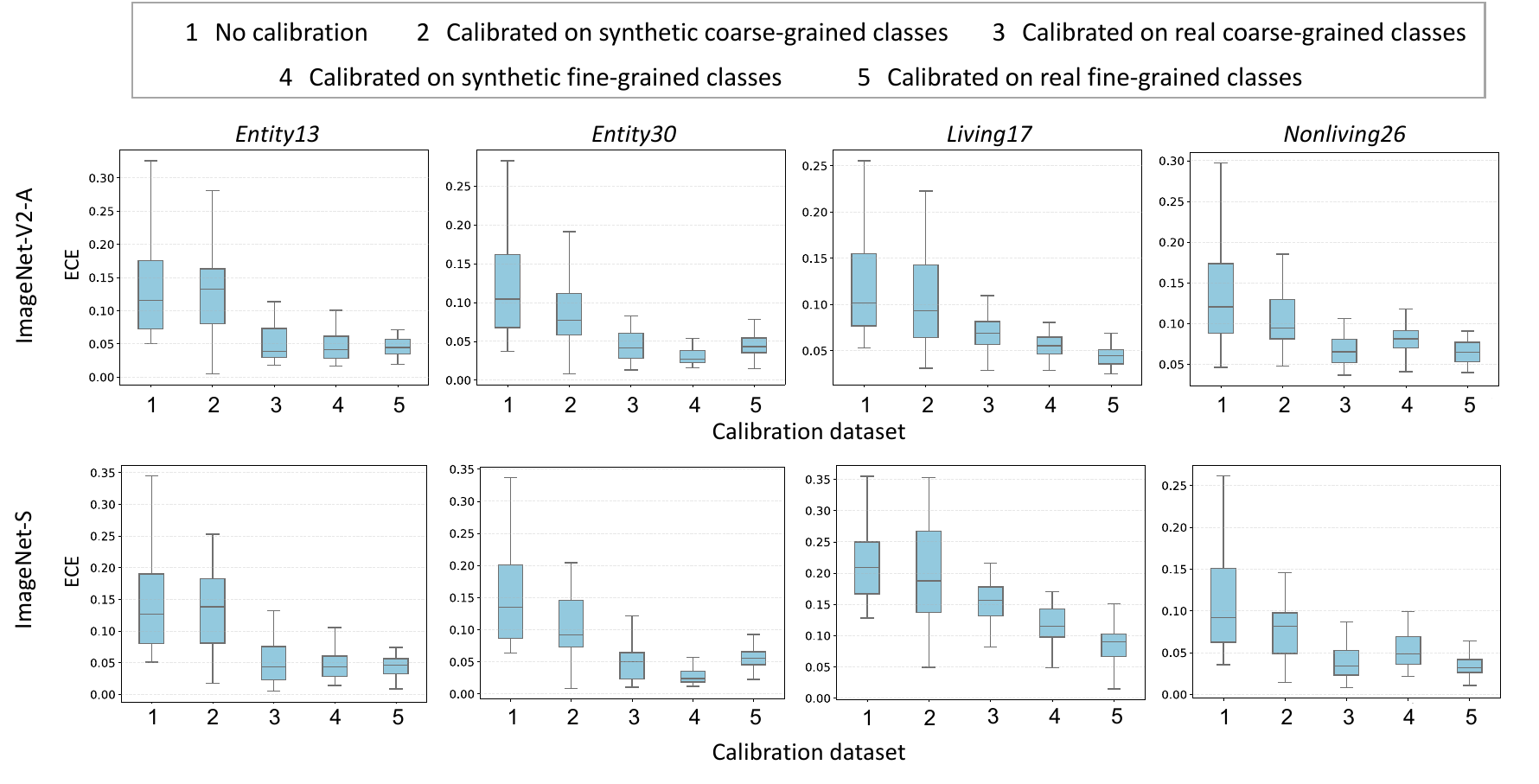}
    \vspace{-20pt}
    \caption{
    \textbf{Can VLMs be calibrated without labeled data? A calibration-by-synthesis approach.}
    Each column represents a pre-defined coarse-fine hierarchical mapping, and each row corresponds to a test set formed according to this class hierarchy.
    The uncertainty estimation performance of VLMs is plotted grouped by different calibration datasets: (1) no calibration; (2) a synthetic dataset generated from coarse-grained class labels; (3) a real dataset with coarse-grained labels; (4) a synthetic dataset generated from fine-grained classes; and (5) a real dataset with fine-grained classes. 
    We observe that calibrating on a synthetic dataset with fine-grained classes is a successful strategy in reducing calibration error and is competitive with the performance of using real data for calibration. 
    This indicates that using a small number of synthetic images, generated from and calibrated with detailed and specific class labels, suffices to calibrate VLMs.
    }
    \label{fig:application}
    \vspace{-5pt}
\end{figure*}

\section{Application: Calibration-by-Synthesis}
\label{sec:application}

In this section, we take some of the calibration robustness findings from \cref{sec:calibration} and demonstrate how they might be applied in a realistic scenario.
We consider the setting where a VLM-based classifier is to be deployed in a new target domain, but labeled data in that domain is unavailable due to reasons of cost, time, or expertise.
Our previous findings give us confidence that such a classifier can still be calibrated even if the calibration set and target domain have a significant distribution shift, if the hierarchy level of the labels differs, and if the calibration set is small.

Our approach is as follows.
First, we construct a set of text descriptions for each class by prompting GPT-3 \cite{brown2020language}, a large language model, following the approach of CuPL \cite{pratt2023does}.
Second, we feed these descriptions to Stable Diffusion \cite{rombach2022high}, a text-to-image model, to synthesize images associated with the descriptions.
These are used as a synthetic, automatically-labeled calibration set.
We consider two situations: (i) the target task is under-specified, so we only have coarsely-defined target classes at calibration time; and (ii) the target task is fully-specified, so we have access to the full (fine-grained) target label set at calibration time.
For the former, the calibration set has $5$ images per coarse class.
For the latter, the calibration set has $1$ image per fine-grained class.
To obtain the coarse--fine label mapping, we use the defined hierarchies (Entity13, Entity30, Living17, and Nonliving26) in the BREEDS benchmark~\cite{santurkar2020breeds} and form the corresponding fine-grained class subsets of ImageNet-V2-A and ImageNet-S for the two target test datasets.

\begin{table}[!t]\centering
\small
\setlength\tabcolsep{1pt}
\begin{tabularx}{\linewidth}{l C C C C C C}
\toprule
& \multicolumn{3}{c}{No calibration} & \multicolumn{3}{c}{Synthetic calibration}\\
\cmidrule(lr){2-4}  \cmidrule(lr){5-7}
Label set & MECE & MAE & $\rho$ & MECE & MAE & $\rho$  \\
		\midrule
		Entity13 &$0.15$& $15.21$ & $0.77$ &$0.05$& $5.07$
               & $0.93$ \\ 
		Entity30 &$0.16$& $16.12$ & $0.75$ &$0.04$& $3.57$
                & $0.94$  \\ 
		Living17  &$0.22$& $21.90$ & $0.74$ &$0.12$& $11.45$
                & $0.94$ \\ 
		Nonliving26  &$0.13$& $12.73$ & $0.79$ &$0.05$& $5.29$
                      & $0.91$ \\
            \cmidrule(lr){1-7}
            Average  &$0.17$& $16.49$ & $0.76$ &$0.07$& $6.35$
                      & $0.93$ \\
            
            \bottomrule
\end{tabularx}
\caption{
\textbf{Uncertainty estimation performance on ImageNet-S for calibration-by-synthesis.}
We report the mean expected calibration error (MECE \revise{$\downarrow$}), the mean absolute error (MAE \revise{$\downarrow$}) of the average confidence with respect to the model accuracy, and Spearman's rank correlation ($\rho$ \revise{$\uparrow$}).
We see that by calibrating on a synthetic set with fine-grained classes, prediction probabilities are informative of the model's performance and its ranking.
}
\vspace{-5pt}
\label{tab:application}
\end{table}

In \cref{fig:application}, we plot the ECE, evaluated on the target dataset with fine-grained class labels, for a set of VLMs grouped by the calibration procedure:
\begin{enumerate*}[label=(\arabic*)]
    \item no calibration;
    \item calibration on a synthetic dataset generated from coarse-grained class labels;
    \item calibration on a real labeled dataset (the ImageNet validation set) with coarse-grained class labels;
    \item calibration on a synthetic dataset generated from fine-grained class labels; and
    \item calibration on a real labeled dataset with fine-grained class labels.
\end{enumerate*} 
Results across four fine-grained label sets show that calibration with synthetic images based on fine-grained classes is remarkably effective, often outperforming calibration with real data. 
For each label set, calibration using fine-grained synthetic classes (Dataset 4) consistently yields lower ECE values compared to uncalibrated models (Dataset 1).
Moreover, coarse synthetic label calibration (Dataset 2) is less effective due to compounded domain shift and label granularity discrepancies.

Beyond model calibration, we showcase the tangible benefits of using calibrated prediction probabilities for choosing models and estimating their accuracy, particularly when labeled data is scarce or absent. \cref{tab:application} assesses how predictive the estimated VLM probabilities are for the model's performance, before and after calibration with a synthetic dataset with fine-grained classes. Calibration improvements are evident: a reduction in both the mean expected calibration error (MECE) and the mean absolute error (MAE) of average confidence in relation to actual model accuracy, along with a boost in rank correlation with model accuracy. Overall, the above analysis shows how our findings about calibration can be applied to a real scenario where labeled calibration data is not available.

\section{Conclusion and Discussion}
In this study, we have investigated the factors that affect the uncertainty estimation performance of Vision--Language Models (VLMs). Our experiments, spanning various facets of uncertainty estimation, have revealed insights into the strengths of VLMs. Notably, when coupled with temperature scaling as a calibration method, VLMs surpass other models in their ability to estimate uncertainty accurately. This finding holds promise for tasks that demand precise uncertainty information. Furthermore, we have demonstrated VLMs' remarkable adaptability---they can be effectively calibrated with datasets of that have different label sets or label hierarchy levels.

Additionally, VLMs exhibit efficiency by maintaining calibration quality with a limited number of images and simplified prompts. Real-world applications confirm their potential, showing that VLMs can be calibrated with a small number of synthetic images.
Looking ahead, we anticipate exciting avenues for research, including exploring alternative calibration methods and investigating uncertainty estimation properties in domains beyond image classification. These insights, we believe, pave the way for more robust and reliable VLMs, contributing to the broader landscape of algorithmic design and multi-modal understanding.

{This work leaves open many interesting directions for future research. We primarily focus on calibrating classification confidence estimates for vanilla/backbone VLM networks. It would be interesting to explore the calibration of regression confidence estimates. Some more parameter-efficient fine-tuning methods for prompts can be included to study the transferability of calibration across prompts. Additionally, a broader class of large language model-based VLMs, such as LLaVA \cite{liu2023llava}, can be further evaluated and examined. These models could potential present a unique calibration property for the lack of inherent capability to generate classification confidence scores.

Lastly, our scope is the calibration of classification confidence estimates for vanilla/backbone VLM networks, rather than the calibration of regression confidence estimates used for other tasks (\eg, object detection and segmentation). In the context of VLMs, these tasks require compound networks with additional modules. We acknowledge that the findings in this study may not transfer to these other tasks, and that this would be interesting to study in future work. 
}

\section*{Acknowledgements}
{We sincerely thank all the anonymous reviewers and area chairs for their constructive comments and valuable suggestions, which have greatly helped in enhancing this paper.}

\section*{Impact Statement}

Poorly calibrated models make under- or over-confident predictions on average, which can have significant negative societal consequences when these models are deployed in the real world.
For example, vehicle safety systems that ascribe a high probability of ``sky'' to a region that actually contains the back of a truck can lead to accidents.
This work investigates what factors might contribute to these calibration errors after temperature scaling, and shows that VLMs are surprisingly robust to a variety of factors when constructing the calibration set.
This increases our confidence in the use of calibrated VLMs in safety-critical sectors like healthcare and autonomous systems.
Nonetheless, cautious deployment is advisable, since there are likely to be other factors that have not been studied here that may contribute to how well-calibrated a VLM is in its target domain.

\bibliography{icml2024/egbib}
\bibliographystyle{icml2024}

\newpage
\appendix
\onecolumn
\section{Appendix.}

\subsection{Datasets:}

\texttt{ImageNet} \cite{imagenet_cvpr09} (\textcolor{blue}{https://www.image-net.org/});\\
\texttt{ImageNet-V2} \cite{shankar2021image} (\textcolor{blue}{https://github.com/modestyachts/ImageNetV2});\\
\texttt{ImageNet-Rendition} \cite{hendrycks2021many} (\textcolor{blue}{https://github.com/hendrycks/robustness}); \\
\texttt{ImageNet-Sketch} \cite{wang2019learning} (\textcolor{blue}{https://github.com/HaohanWang/ImageNet-Sketch});\\
\texttt{ObjectNet} \cite{barbu2019objectnet} (\textcolor{blue}{https://objectnet.dev/download.html});\\
\texttt{CIFAR-10} \cite{krizhevsky2009learning}: (\textcolor{blue}{https://www.cs.toronto.edu/~kriz/cifar.html});\\
\texttt{CIFAR-10.1} \cite{recht2018cifar}: (\textcolor{blue}{https://github.com/modestyachts/CIFAR-10.1});\\
\texttt{CIFAR-10.2} \cite{recht2018cifar}: (\textcolor{blue}{https://github.com/modestyachts/CIFAR-10.1});\\
\texttt{CINIC} \cite{darlow2018cinic}: (\textcolor{blue}{https://www.v7labs.com/open-datasets/cinic-10});\\
\texttt{DomainNet} \cite{peng2019moment}: (\textcolor{blue}{http://ai.bu.edu/M3SDA/});\\

\subsection{Models Included in Experiments}

\textbf{(1) Vision--language models:}

(1) we use the zero-shot CLIP models provided in OpenCLIP \cite{ilharco_gabriel_2021_5143773}. They are listed as follows in the pattern (\texttt{architecture}, \texttt{source}):

(\texttt{RN50},~\texttt{openai})

(\texttt{RN50},~\texttt{yfcc15m})

(\texttt{RN50},~\texttt{cc12m})

(\texttt{ViT-B-32},~\texttt{openai})

(\texttt{ViT-B-32},~\texttt{laion400m\_e32})

(\texttt{ViT-B-32},~\texttt{laion2b\_s34b\_b79k})

(\texttt{ViT-B-16},~\texttt{openai})

(\texttt{ViT-B-16},~\texttt{laion400m\_e32})

(\texttt{ViT-L-14},~\texttt{openai})

(\texttt{ViT-L-14},~\texttt{laion400m\_e32})

(\texttt{ViT-H-14},~\texttt{laion2b\_s32b\_b79k})

(\texttt{ViT-g-14},~\texttt{laion2b\_s34b\_b88k})

(\texttt{ViT-bigG-14},~\texttt{laion2b\_s39b\_b160k})

(\texttt{convnext\_base},~\texttt{laion400m\_s13b\_b51k})

(\texttt{convnext\_base\_w},~\texttt{laion\_aesthetic\_s13b\_b82k})

(\texttt{convnext\_xxlarge},~\texttt{laion2b\_s34b\_b82k\_augreg})

(\texttt{ViT-B-32},~\texttt{Model-B-32\_Data-80M\_Samples-34B\_lr-1e-3\_bs-88k.pt})

(\texttt{ViT-B-16},~\texttt{Model-B-16\_Data-80M\_Samples-34B\_lr-1e-3\_bs-88k.pt})

(\texttt{ViT-L-14},~\texttt{Model-L-14\_Data-80M\_Samples-34B\_lr-1e-3\_bs-88k.pt})

(\texttt{ViT-B-32},~\texttt{datacomp\_m\_s128m\_b4k})

(\texttt{ViT-B-32},~\texttt{datacomp\_s\_s13m\_b4k})

(\texttt{ViT-B-16},~\texttt{datacomp\_l\_s1b\_b8k})

(\texttt{ViT-L-14},~\texttt{datacomp\_xl\_s13b\_b90k})

(2) Other vision--language models: 

(\texttt{EVA02-B-16},~\texttt{merged2b\_s8b\_b131k})

(\texttt{EVA02-L-14},~\texttt{merged2b\_s8b\_b131k})

(\texttt{ViT-B-32-quickgelu},~\texttt{metaclip\_400m})

(\texttt{ViT-B-16-quickgelu},~\texttt{metaclip\_fullcc})

(\texttt{ViT-B-16-SigLIP},~\texttt{webli})

(\texttt{ViT-B-16-SigLIP-256},~\texttt{webli})

(\texttt{ViT-H-14-CLIPA},~\texttt{datacomp1b})

(\texttt{ViT-L-14-CLIPA},~\texttt{datacomp1b})

Flava can be derived from Hugging Face, Transformer module \cite{wolf-etal-2020-transformers}.

BLIP \cite{li2022blip}, BLIP-2 \cite{li2023blip} and ALBEF \cite{li2021align} models are publicly available through LAVIS package \cite{li-etal-2023-lavis}.

\textbf{(2) ImageNet-trained models:}

we compare VLMs' uncertainty estimate performance with following models from TIMM \cite{rw2019timm}

Trained from scratch on ImageNet:

\texttt{swin\_base\_patch4\_window12\_384}

\texttt{deit\_small\_distilled\_patch16\_224}

\texttt{deit\_base\_patch16\_224}

\texttt{swin\_small\_patch4\_window7\_224}

\texttt{deit\_base\_patch16\_384}

\texttt{wide\_resnet50\_2}

\texttt{convnext\_base}

\texttt{tv\_resnet50}

\texttt{densenet121}

\texttt{inception\_v4}

\texttt{resmlp\_36\_224}

\texttt{xception}

\texttt{vgg19}

Pre-trained on a larger dataset then fine-tuned on ImageNet training split:

\texttt{vit\_base\_patch16\_224.orig\_in21k\_ft\_in1k}

\texttt{vit\_base\_patch32\_224}

\texttt{vit\_large\_r50\_s32\_384}

\texttt{beit\_large\_patch16\_224}

\texttt{vit\_large\_patch16\_384}

\texttt{swin\_large\_patch4\_window12\_384}

\texttt{convnext\_small.in12k\_ft\_in1k}

\texttt{ig\_resnext101\_32x16d}

\texttt{mixer\_b16\_224\_miil}

\texttt{resnetv2\_50x1\_bitm}

\texttt{ig\_resnext101\_32x8d}

\subsection{Prompt Template}
In this section, we provide the prompt set used in our experiments. For experiments included in \cref{sec:crosstask}, \cref{sec:hierarchy}, \cref{sec:distance} and \cref{sec:methods}, we also use default prompt sets provided by \citet{radford2021learning}. For experiments in \cref{sec:prompt}, the calibration prompt set is ``\textit{a photo of a $<$class$>$}''. In \cref{sec:application}, the calibration prompt set is ``\textit{a photo of a $<$class$>$}'', while the test prompt set is generated by CuPL.

\subsection{Computation Resources} 
PyTorch version is \textcolor{black}{1.10.0+cu111} and timm version is \textcolor{black}{0.8.21dev0}. 
All experiment is run on one 3090 and the CPU \textcolor{black}{AMD EPYC 7343 16-Core Processor}.

\newpage

\subsection{Experiments on Specialized Domain: Satellite Images}

We further conduct experiments on whether our findings maintain on more specialized domains: Satellite images. We use EuroSAT \cite{helber2019eurosat} as the target set. Two different calibration datasets are considered: 1) ImageNet validation set; 2) A generated dataset which consists of $68$ images from finer-grained classes of Living17 hierarchy. The calibration method is temperature scaling. We see that our observations persist on such a specialized domain that VLMs can be calibrated across different label sets using a few samples.

\begin{figure}[!ht]
    \centering
    \vspace{-10pt}
    \includegraphics[width=\linewidth]{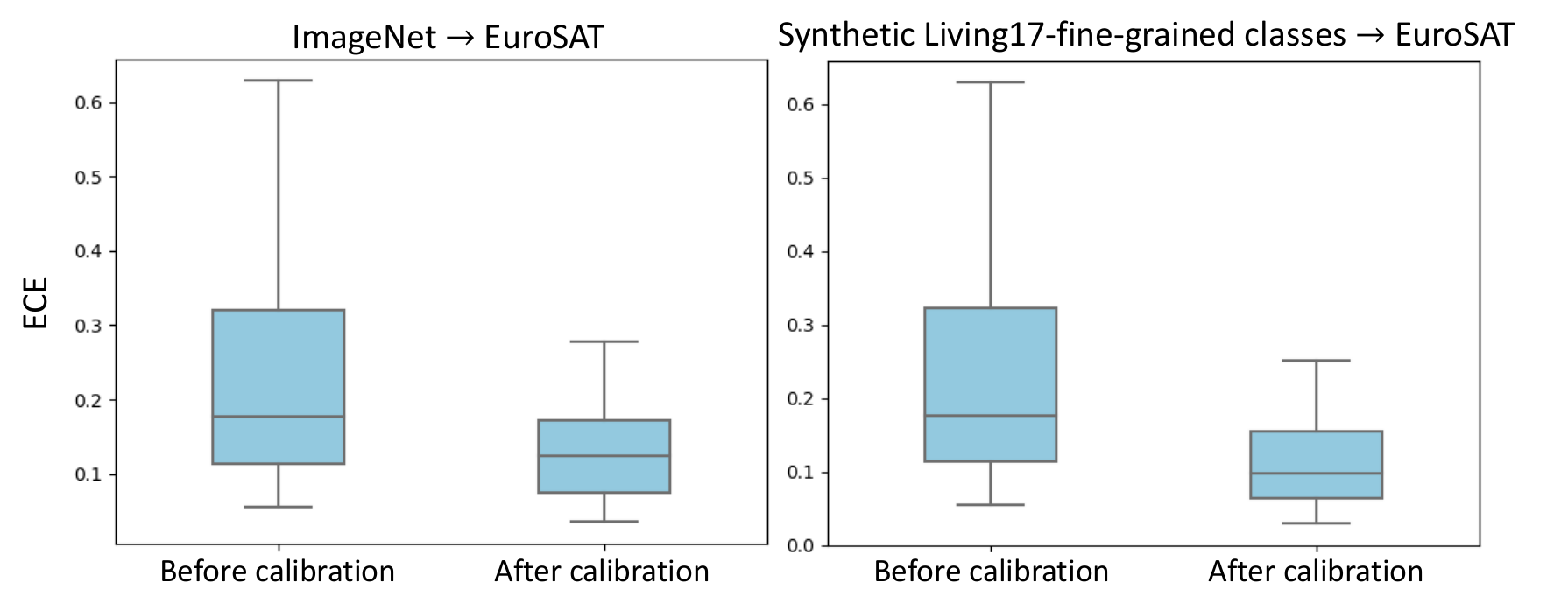}
    \vspace{-20pt}

    \caption{
    \textbf{Uncertainty estimate performance of VLMs when the target domain contains Satellite images.}
    Left: The calibration set is ImageNet-Validation set. Right: The calibration set is a synthesized dataset containing fine-grained classes of Living17.
    }
    
    \label{fig:euro}
    \vspace{-10pt}
\end{figure}

\subsection{Additional Results for Adaptability to Different Calibration Label Sets}

Following the same practice as \cref{sec:crosstask}, we additionally validate that the findings on ImageNet-S and ImageNet-R persist when the target datasets change to ImageNet-A or ObjectNet.

\begin{figure}[!ht]
    \centering
    \includegraphics[width=\linewidth]{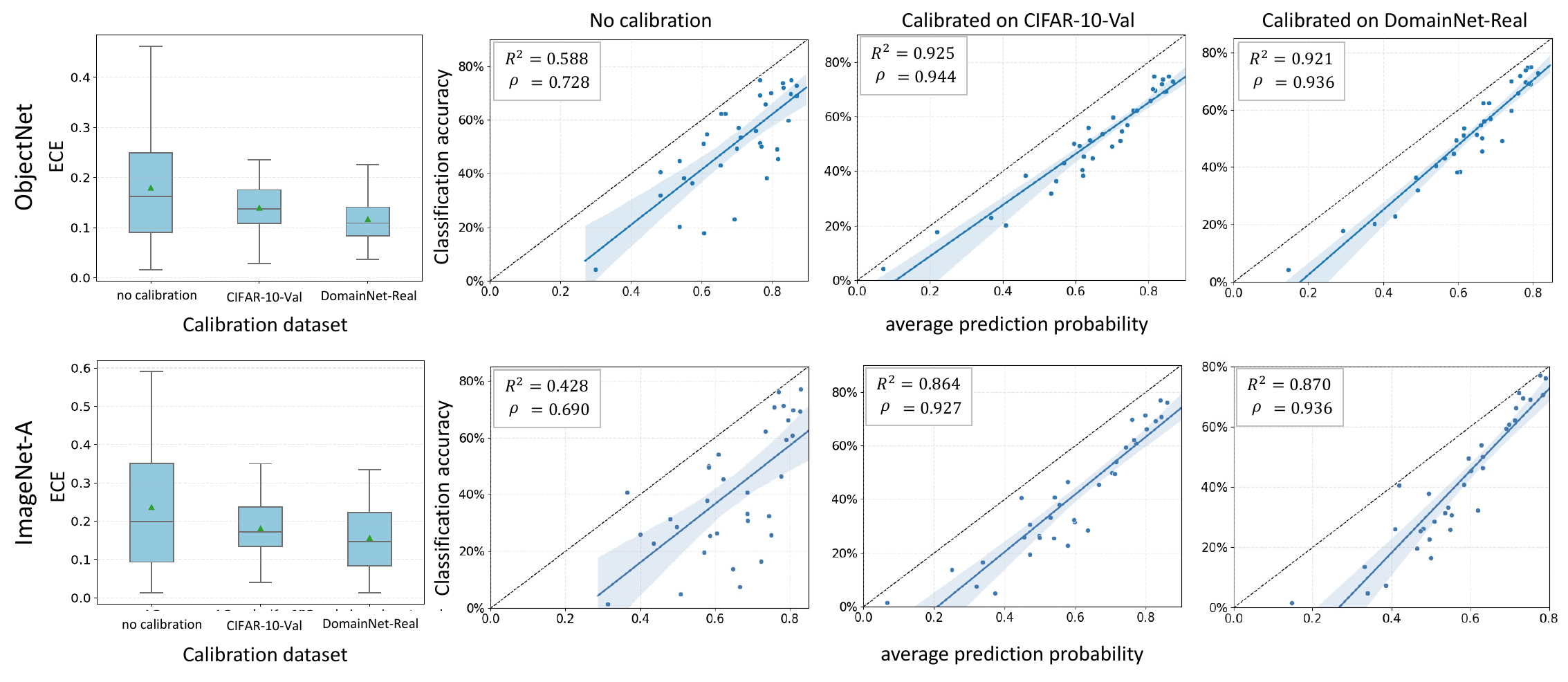}
    \vspace{-20pt}
    \caption{
    \textbf{Adaptability of VLMs to different calibration label sets.}
    \textbf{Left: Calibration error reduction.}
    Here, we observe a significant decrease in the expected calibration error for VLMs following cross-label-set calibration, as opposed to when no calibration is applied.
    \textbf{Right: Correlation between VLM prediction probability and classification accuracy.}
    This graph illustrates the classification accuracy of VLMs on ObejctNet and ImageNet-A against their average prediction probability, before and after calibration with CIFAR-10-Val or DomainNet-Real. Each point represents a model, with the dashed black line indicating perfect calibration ($y=x$). The data showcases a strong linear and rank correlation, even when models are calibrated on label sets different from the target, proving the effectiveness of cross-label-set calibration for VLMs.
    }
    \label{fig:supp_crosstask}
    \vspace{-10pt}
\end{figure}

\newpage
\subsection{Additional Results for Calibration Across Semantic Hierarchy Levels}

We also show the ECE of VLMs calibrated with label hierarchies differing in granularity from the target dataset (ImageNet-V2-A). The results validate that VLMs can be calibrated across label hierarchies. 

\begin{figure}[!ht]\centering
\includegraphics[width=0.6\linewidth]{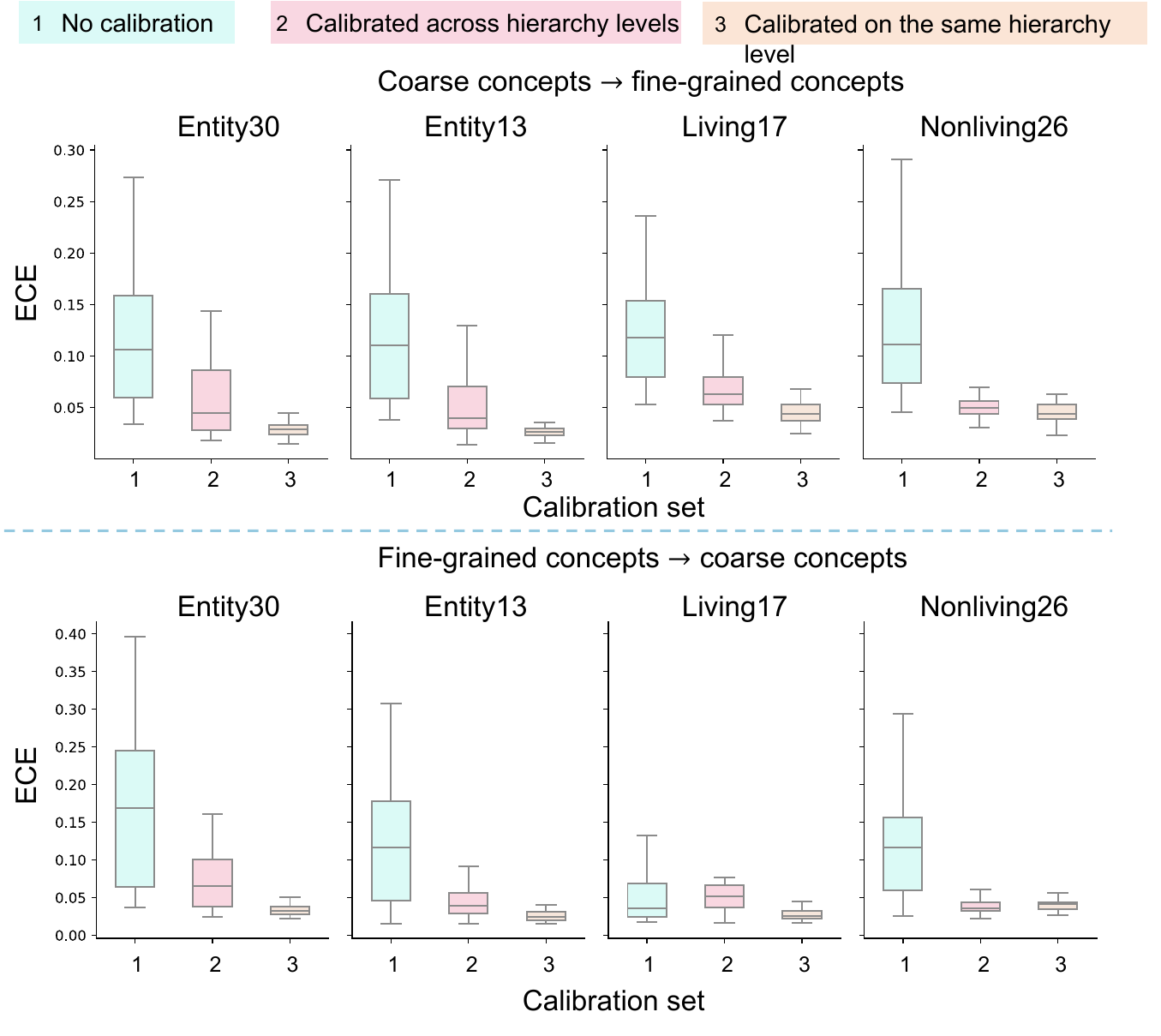}
\vspace{-15pt}
\caption{
\textbf{Robustness of VLM calibration to label hierarchy levels.}
This figure presents box plots summarizing the calibration errors (ECEs) of VLMs calibrated with label hierarchies differing in granularity from the target dataset (ImageNet-V2-A). The top row shows calibration at a coarser level, and the bottom row at a finer level. Despite not matching the calibration precision of same-level calibration, the minimal differences indicate the robustness of VLM calibration to label granularity.
}
\label{fig:breeds_in-v2-a}
\end{figure}

\subsection{Transferability of Calibration Across Prompts Using Advanced Prompt Tuning Methods}
We additionally test whether our observations that VLMs can be calibrated by using ``\textit{a photo of a $<$class$>$}'' persist when the test prompt set is obtained via advanced prompt fine-tuning methods, such as MaPLe \cite{khattak2023maple} or PromptSRC~\cite{Khattak_2023_ICCV}. We use ViT-B/16 as the visual backbone and fine-tune them on ImageNet validation with $16$ shots per class. Following the procedure in \ref{sec:prompt}, we randomly selected a single prompt for calibration and evaluated whether the searched temperature scalar is useful on ImageNet-Sketch, ImageNet-V2 and ImageNet-Rendition for the prompt-tuned models. The results are summarized in . We have a consistent observation with \ref{fig:prompt} that VLM can be calibrated with one single prompt.

\begin{table}[!ht]\centering
\small
\setlength\tabcolsep{1pt}
\begin{tabularx}{\linewidth}{l C C C C C C}
\toprule
 & \multicolumn{2}{c}{ImageNet-S} & \multicolumn{2}{c}{ImageNet-V2-A} & \multicolumn{2}{c}{ImageNet-R} \\
\cmidrule(lr){2-3}  \cmidrule(lr){4-5} \cmidrule(lr){6-7}
  & Before & After &  Before & After &  Before & After  \\
		\midrule
		MaPLe &$0.0364$& $0.0073$ & $0.0159$ &$0.0165$& $0.0264$
               & $0.0073$ \\ 
		PromptSRC &$0.0323$& $0.0071$ & $0.0175$ &$0.0173$& $0.0328$
                & $0.0071$  \\ 
            \bottomrule
\end{tabularx}
\caption{
\textbf{Uncertainty estimation performance on ImageNet-S, ImageNet-V2-A and ImageNet-R for cross prompt set calibration.}
We report the expected calibration error (ECE \revise{$\downarrow$}) before and after calibration on ImageNet validation set.
We see that by calibrating on one single prompt on ImageNet-Val with temperature scaling, VLMs become well-calibrated.
}
\label{tab:application}
\end{table}

\end{document}